# Higher Order Conditional Random Fields in Deep Neural Networks

Anurag Arnab, Sadeep Jayasumana, Shuai Zheng, Philip H.S. Torr

University of Oxford
{firstname.lastname}@eng.ox.ac.uk

**Abstract.** We address the problem of semantic segmentation using deep learning. Most segmentation systems include a Conditional Random Field (CRF) to produce a structured output that is consistent with the image's visual features. Recent deep learning approaches have incorporated CRFs into Convolutional Neural Networks (CNNs), with some even training the CRF end-to-end with the rest of the network. However, these approaches have not employed higher order potentials, which have previously been shown to significantly improve segmentation performance. In this paper, we demonstrate that two types of higher order potential, based on object detections and superpixels, can be included in a CRF embedded within a deep network. We design these higher order potentials to allow inference with the differentiable mean field algorithm. As a result, all the parameters of our richer CRF model can be learned end-to-end with our pixelwise CNN classifier. We achieve state-of-the-art segmentation performance on the PASCAL VOC benchmark with these trainable higher order potentials.

## 1 Introduction

Semantic segmentation involves assigning a visual object class label to every pixel in an image, resulting in a segmentation with a semantic meaning for each segment. While a strong pixel-level classifier is critical for obtaining high accuracy in this task, it is also important to enforce the consistency of the semantic segmentation output with visual features of the image. For example, segmentation boundaries should usually coincide with strong edges in the image, and regions in the image with similar appearance should have the same label.

Recent advances in deep learning have enabled researchers to create stronger classifiers, with automatically learned features, within a Convolutional Neural Network (CNN) [1–3]. This has resulted in large improvements in semantic segmentation accuracy on widely used benchmarks such as PASCAL VOC [4]. CNN classifiers are now considered the standard choice for pixel-level classifiers used in semantic segmentation.

On the other hand, probabilistic graphical models have long been popular for structured prediction of labels, with constraints enforcing label consistency. Conditional Random Fields (CRFs) have been the most common framework, and various rich and expressive models [5–7], based on higher order clique potentials, have been developed to improve segmentation performance.

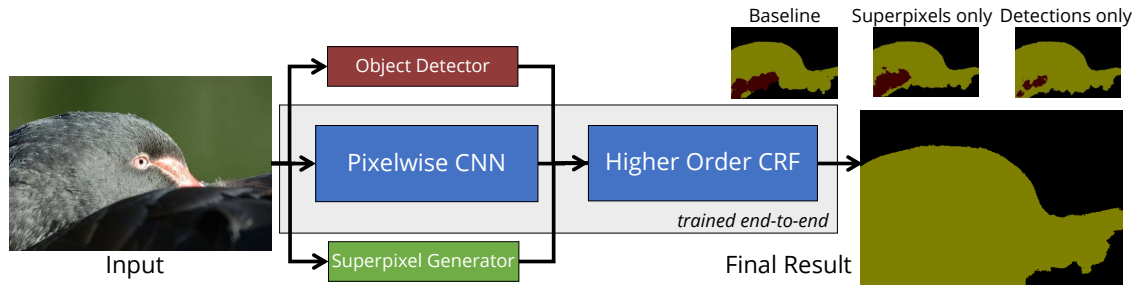

Fig. 1: **Overview of our system** We train a Higher Order CRF end-to-end with a pixelwise CNN classifier. Our higher order detection and superpixel potentials improve significantly over our baseline containing only pairwise potentials.

Whilst some deep learning methods showed impressive performance in semantic segmentation without incorporating graphical models [3, 8], current state-of-the-art methods [9–12] have all incorporated graphical models into the deep learning framework in some form. However, we observe that the CRFs that have been incorporated into deep learning techniques are still rather rudimentary as they consist of only unary and pairwise potentials [10]. In this paper, we show that CRFs with carefully designed higher order potentials (potentials defined over cliques consisting of more than two nodes) can also be modelled as CNN layers when using mean field inference [13]. The advantage of performing CRF inference within a CNN is that it enables joint optimisation of CNN classifier weights and CRF parameters during the end-to-end training of the complete system. Intuitively, the classifier and the graphical model learn to optimally co-operate with each other during the joint training.

We introduce two types of higher order potential into the CRF embedded in our deep network: object-detection based potentials and superpixel-based potentials. The primary idea of using object-detection potentials is to use the outputs of an off-the-shelf object detector as additional semantic cues for finding the segmentation of an image. Intuitively, an object detector with a high recall can help the semantic segmentation algorithm by finding objects appearing in an image. As shown in Fig. 1, our method is able to recover from poor segmentation unaries when we have a confident detector response. However, our method is robust to false positives identified by the object detector since CRF inference identifies and rejects false detections that do not agree with other types of energies present in the CRF.

Superpixel-based higher order potentials encourage label consistency over superpixels obtained by oversegmentation. This is motivated by the fact that regions defined by superpixels are likely to contain pixels from the same visual object. Once again, our formulation is robust to the violations of this assumption and errors in the initial superpixel generation step. In practice, we noted that this potential is effective for getting rid of small regions of spurious labels that are inconsistent with the correct labels of their neighbours.

We evaluate our higher order potentials on the PASCAL VOC 2012 semantic segmentation benchmark as well as the PASCAL Context dataset, to show significant improvements over our baseline and achieve state-of-the art results.

## 2   Related Work

Before deep learning became prominent, semantic segmentation was performed with dense hand-crafted features which were fed into a per-pixel or region classifier [14]. The individual predictions made by these classifiers were often noisy as they lacked global context, and were thus post-processed with a CRF, making use of prior knowledge such as the fact that nearby pixels, as well as pixels of similar appearance, are likely to share the same class label [14, 15].

The CRF model of [14] initially contained only unary and pairwise terms in an 8-neighbourhood, which [16] showed can result in shrinkage bias. Numerous improvements to this model were subsequently proposed including: densely connected pairwise potentials facilitating interactions between all pairs of image pixels [17], formulating higher order potentials defined over cliques larger than two nodes [5, 16] in order to capture more context, modelling co-occurrence of object classes [18–20], and utilising the results of object detectors [6, 21, 22].

Recent advances in deep learning have allowed us to replace hand-crafted features with features learned specifically for semantic segmentation. The strength of these representations was illustrated by [3] who achieved significant improvements over previous hand-crafted methods without using any CRF post-processing. Chen *et al.* [12] showed further improvements by post-processing the results of a CNN with a CRF. Subsequent works [9–11, 23] have taken this idea further by incorporating a CRF as layers within a deep network and then learning parameters of both the CRF and CNN together via backpropagation.

In terms of enhancements to conventional CRF models, Ladicky *et al.* [6] proposed using an off-the-shelf object detector to provide additional cues for semantic segmentation. Unlike other approaches that refine a bounding-box detection to produce a segmentation [8, 24], this method used detector outputs as a soft constraint and can thus recover from object detection errors. Their formulation, however, used graph-cut inference, which was only tractable due to the absence of dense pairwise potentials. Object detectors have also been used by [21, 25], who also modelled variables that describe the degree to which an object hypothesis is accepted.

We formulate the detection potential in a different manner to [6, 21, 25] so that it is amenable to mean field inference. Mean field permits inference with dense pairwise connections, which results in substantial accuracy improvements [10, 12, 17]. Furthermore, mean field updates related to our potentials are differentiable and its parameters can thus be learned in our end-to-end trainable architecture.

We also note that while the semantic segmentation problem has mostly been formulated in terms of pixels [3, 10, 14], some have expressed it in terms of superpixels [26–28]. Superpixels can capture more context than a single pixel and computational costs can also be reduced if one considers pairwise interactions between superpixels rather than individual pixels [21]. However, such superpixel representations assume that the segments share boundaries with objects in an image, which is not always true. As a result, several authors [5, 7] have employed higher order potentials defined over superpixels that encourage label consistency over regions, but do not strictly enforce it. This approach also allows multiple, non-hierarchical layers of superpixels to be integrated. Our formulation uses this kind of higher order potential, but in an end-to-end trainable CNN.

Graphical models have been used with CNNs in other areas besides semantic segmentation, such as in pose-estimation [29] and group activity recognition [30]. Alternatively, Ionescu *et al.* [31] incorporated structure into a deep network with structured matrix layers and matrix backpropagation. However, the nature of models used in these works is substantially different to ours. Some early works that advocated gradient backpropagation through graphical model inference for parameter optimisation include [32, 33] and [34].

Our work differentiates from the above works since, to our knowledge, we are the first to propose and conduct a thorough experimental investigation of higher order potentials that are based on detection outputs and superpixel segmentation in a CRF which is learned end-to-end in a deep network. Note that although [7] formulated mean field inference with higher order potentials, they did not consider object detection potentials at all, nor were the parameters learned.

## 3    Conditional Random Fields

We now review conditional random fields used in semantic segmentation and introduce the notation used in the paper. Take an image $\mathbf{I}$ with $N$ pixels, indexed $1, 2, \ldots, N$. In semantic segmentation, we attempt to assign every pixel a label from a predefined set of labels $\mathcal{L} = \{l_1, l_2, \ldots, l_L\}$. Define a set of random variables $X_1, X_2, \ldots, X_N$, one for each pixel, where each $X_i \in \mathcal{L}$. Let $\mathbf{X} = [X_1 \ X_2 \ \ldots \ X_N]^T$. Any particular assignment $\mathbf{x}$ to $\mathbf{X}$ is thus a solution to the semantic segmentation problem.

We use notations $\{\mathbf{V}\}$, and $\mathbf{V}^{(i)}$ to represent the set of elements of a vector $\mathbf{V}$, and the $i^{\text{th}}$ element of $\mathbf{V}$, respectively. Given a graph $G$ where the vertices are from $\{\mathbf{X}\}$ and the edges define connections among these variables, the pair $(\mathbf{I}, \mathbf{X})$ is modelled as a CRF characterised by $\Pr(\mathbf{X} = \mathbf{x}|\mathbf{I}) = (1/Z(\mathbf{I})) \exp(-E(\mathbf{x}|\mathbf{I}))$, where $E(\mathbf{x}|\mathbf{I})$ is the *energy* of the assignment $\mathbf{x}$ and $Z(\mathbf{I})$ is the normalisation factor known as the partition function. We drop the conditioning on $\mathbf{I}$ hereafter to keep the notation uncluttered. The energy $E(\mathbf{x})$ of an assignment is defined using the set of cliques $\mathcal{C}$ in the graph $G$. More specifically, $E(\mathbf{x}) = \sum_{c \in \mathcal{C}} \psi_c(\mathbf{x}_c)$, where $\mathbf{x}_c$ is a vector formed by selecting elements of $\mathbf{x}$ that correspond to random variables belonging to the clique $c$, and $\psi_c(.)$ is the cost function for the clique $c$. The function, $\psi_c(.)$, usually uses prior knowledge about a good segmentation, as well as information from the image, the observation the CRF is conditioned on.

Minimising the energy yields the maximum a posteriori (MAP) labelling of the image *i.e.* the most probable label assignment given the observation (image). When dense pairwise potentials are used in the CRF to obtain higher accuracy, exact inference is impracticable, and one has to resort to an approximate inference method such as mean field inference [17]. Mean field inference is particularly appealing in a deep learning setting since it is possible to formulate it as a Recurrent Neural Network [10].

## 4  CRF with Higher Order Potentials

Many CRF models that have been incorporated into deep learning frameworks [10, 12] have so far used only unary and pairwise potentials. However, potentials defined on higher order cliques have been shown to be useful in previous works such as [7, 16]. The key contribution of this paper is to show that a number of explicit higher order potentials can be added to CRFs to improve image segmentation, while staying compatible with deep learning. We formulate these higher order potentials in a manner that mean field inference can still be used to solve the CRF. Advantages of mean field inference are twofold: First, it enables efficient inference when using densely-connected pairwise potentials. Multiple works, [10, 33] have shown that dense pairwise connections result in substantial accuracy improvements, particularly at image boundaries [12, 17]. Secondly, we keep all our mean field updates differentiable with respect to their inputs as well as the CRF parameters introduced. This design enables us to use backpropagation to automatically learn all the parameters in the introduced potentials.

We use two types of higher order potential, one based on object detections and the other based on superpixels. These are detailed in Sections 4.1 and 4.2 respectively. Our complete CRF model is represented by

$$E(\mathbf{x}) = \sum_i \psi_i^U(x_i) + \sum_{i<j} \psi_{ij}^P(x_i, x_j) \ + \sum_d \psi_d^{\text{Det}}(\mathbf{x}_d) + \sum_s \psi_s^{\text{SP}}(\mathbf{x}_s), \quad (1)$$

where the first two terms $\psi_i^U(.)$ and $\psi_{ij}^P(.,.)$ are the usual unary and densely-connected pairwise energies [17] and the last two terms are the newly introduced higher order energies. Energies from the object detection take the form $\psi_d^{\text{Det}}(\mathbf{x}_d)$, where vector $\mathbf{x}_d$ is formed by elements of $\mathbf{x}$ that correspond to the foreground pixels of the $d^{\text{th}}$ object detection. Superpixel label consistency based energies take the form $\psi_s^{\text{SP}}(\mathbf{x}_s)$, where $\mathbf{x}_s$ is formed by elements of $\mathbf{x}$ that correspond to the pixels belonging to the $s^{\text{th}}$ superpixel.

### 4.1  Object Detection Based Potentials

Semantic segmentation errors can be classified into two broad categories [35]: recognition and boundary errors. Boundary errors occur when semantic labels are incorrect at the edges of objects, and it has been shown that densely connected CRFs with appearance-consistency terms are effective at combating this problem [17]. On the other hand, recognition errors occur when object categories are recognised incorrectly or not at all. A CRF with only unary and pairwise potentials cannot effectively correct these errors since they are caused by poor unary classification. However, we propose that a state-of-the-art object detector [36, 37] capable of recognising and localising objects, can provide important information in this situation and help reduce the recognition error, as shown in Fig. 2.

A key challenge in feeding-in object-detection potentials to semantic segmentation are false detections. A naïve approach of adding an object detector's output to a CRF formulated to solve the problem of semantic segmentation would confuse the CRF due to the presence of the false positives in the detector's output. Therefore, a robust formulation, which can automatically reject object detection false positives when they do not agree with other types of potentials in the CRF, is desired. Furthermore, since we are aiming for an end-to-end trainable CRF which can be incorporated into a deep neural network, the energy formulation should permit a fully differentiable inference procedure. We now propose a formulation which has both of these desired properties.

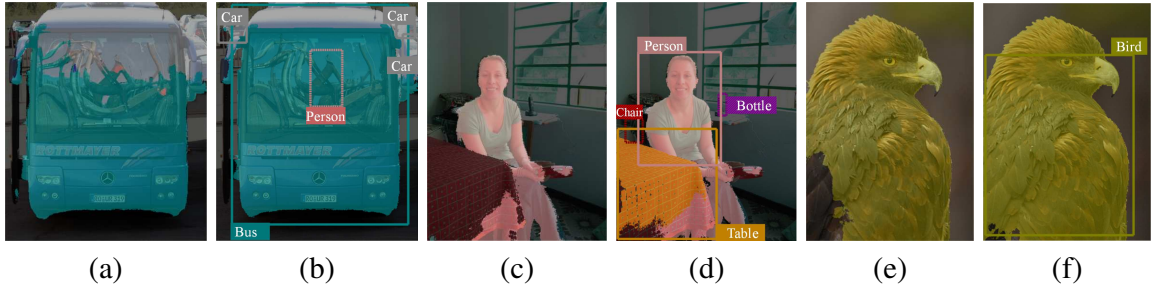

|  (a)  |  (b)  |  (c)  |  (d)  |  (e)  |  (f)  |

Fig. 2: **Utility of object detections as another cue for semantic segmentation** For every pair, segmentation on the left was produced with only unary and pairwise potentials. Detection based potentials were added to produce the result on the right. Note how we are able to improve our segmentations for the bus, table and bird over their respective baselines. Furthermore, our system is able to reject erroneous detections such as the person in (b) and the bottle and chair in (d). Images were taken from the PASCAL VOC 2012 reduced validation set. Baseline results were produced using the public code and model of [10].

Assume that we have $D$ object detections for a given image, and that the $d^{\text{th}}$ detection is of the form $(l_d, s_d, F_d)$, where $l_d \in \mathcal{L}$ is the class label of the detected object, $s_d$ is the confidence score of the detection, and $F_d \subseteq \{1, 2, \ldots, N\}$, is the set of indices of the pixels belonging to the foreground of the detection. The foreground within a detection bounding box could be obtained using a foreground/background segmentation method (*i.e.* GrabCut [38]), and represents a crude segmentation of the detected object. Using our detection potentials, we aim to encourage the set of pixels represented by $F_d$, to take the label $l_d$. However, this should not be a hard constraint since the foreground segmentation could be inaccurate and the detection itself could be a false detection. We therefore seek a soft constraint that assigns a penalty if a pixel in $F_d$ takes a label other than $l_d$. Moreover, if other energies used in the CRF strongly suggest that many pixels in $F_d$ do not belong to the class $l_d$, the detection $d$ should be identified as invalid.

An approach to accomplish this is described in [6] and [21]. However, in both cases, dense pairwise connections were absent and different inference methods were used. In contrast, we would like to use the mean field approximation to enable efficient inference with dense pairwise connections [17], and also because its inference procedure is fully differentiable. We therefore use a detection potential formulation quite different to the ones used in [6] and [21].

In our formulation, as done in [6] and [21], we first introduce latent binary random variables $Y_1, Y_2, \ldots Y_D$, one for each detection. The interpretation for the random variable $Y_d$ that corresponds to the $d^{\text{th}}$ detection is as follows: If the $d^{\text{th}}$ detection has been found to be valid after inference, $Y_d$ will be set to 1, it will be 0 otherwise. Mean field inference probabilistically decides the final value of $Y_d$. Note that, through this formulation, we can account for the fact that the initial detection could have been a false positive: some of the detections obtained from the object detector may be identified to be false following CRF inference.

All $Y_d$ variables are added to the CRF which previously contained only $X_i$ variables. Let each $(\mathbf{X}_d, Y_d)$, where $\{\mathbf{X}_d\} = \{X_i \in \{\mathbf{X}\} | i \in F_d\}$, form a clique $c_d$ in the CRF. We define the detection-based higher order energy associated with a particular assignment $(\mathbf{x}_d, y_d)$ to the clique $(\mathbf{X}_d, Y_d)$ as follows:

$$\psi_d^{\mathrm{Det}}(\mathbf{X}_d = \mathbf{x}_d, Y_d = y_d) = \begin{cases} w_{\mathrm{Det}} \frac{s_d}{n_d} \sum_{i=1}^{n_d} [x_d^{(i)} = l_d] & \text{if } y_d = 0, \\ \\ w_{\mathrm{Det}} \frac{s_d}{n_d} \sum_{i=1}^{n_d} [x_d^{(i)} \neq l_d] & \text{if } y_d = 1, \end{cases} \tag{2}$$

where $n_d = |F_d|$ is the number of foreground pixels in the $d^{\mathrm{th}}$ detection, $x_d^{(i)}$ is the $i^{\mathrm{th}}$ element of the vector $\mathbf{x}_d$, $w_{\mathrm{Det}}$ is a learnable weight parameter, and $[\,.\,]$ is the Iverson bracket. Note that this potential encourages $X_d^{(i)}$s to take the value $l_d$ when $Y_d$ is 1, and at the same time encourages $Y_d$ to be 0 when many $X_d^{(i)}$s do not take $l_d$. In other words, it enforces the consistency among $X_d^{(i)}$s and $Y_d$.

An important property of the above definition of $\psi_d^{\mathrm{Det}}(.)$ is that it can be simplified as a sum of pairwise potentials between $Y_d$ and each $X_d^{(i)}$ for $i = 1, 2, \ldots, n_d$. That is,

$$\psi_d^{\mathrm{Det}}(\mathbf{X}_d = \mathbf{x}_d, Y_d = y_d) = \sum_{i=1}^{n_d} f_d(x_d^{(i)}, y_d), \text{ where,}$$

$$f_d(x_d^{(i)}, y_d) = \begin{cases} w_{\mathrm{Det}} \frac{s_d}{n_d} [x_d^{(i)} = l_d] & \text{if } y_d = 0, \\ \\ w_{\mathrm{Det}} \frac{s_d}{n_d} [x_d^{(i)} \neq l_d] & \text{if } y_d = 1. \end{cases} \tag{3}$$

We make use of this simplification in Section 5 when deriving the mean field updates associated with this potential.

For the latent $Y$ variables, in addition to the joint potentials with $X$ variables, described in Eq. (2) and (3), we also include unary potentials, which are initialised from the score $s_d$ of the object detection. The underlying idea is that if the object detector detects an object with high confidence, the CRF in turn starts with a high initial confidence about the validity of that detection. This confidence can, of course, change during the CRF inference depending on other information (*e.g.* segmentation unary potentials) available to the CRF.

Examples of input images with multiple detections and GrabCut foreground masks are shown in Figure 3. Note how false detections are ignored and erroneous parts of the foreground mask are also largely ignored.

## 4.2  Superpixel Based Potentials

The next type of higher order potential we use is based on the idea that superpixels obtained from oversegmentation [39, 40] quite often contain pixels from the same visual object. It is therefore natural to encourage pixels inside a superpixel to have the same semantic label. Once again, this should not be a hard constraint in order to keep the algorithm robust to initial superpixel segmentation errors and to violations of this key assumption.

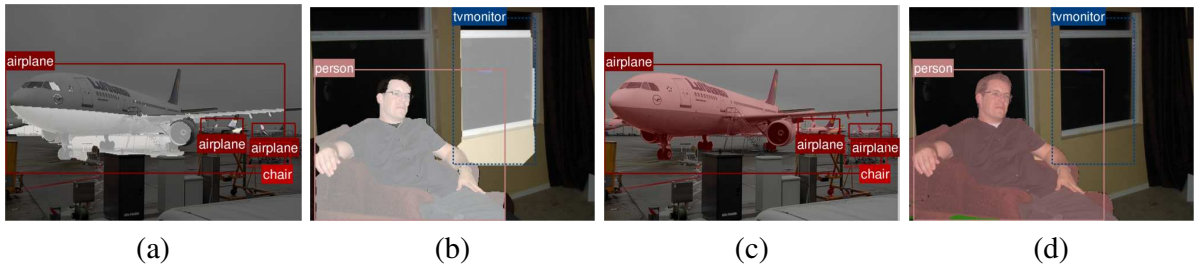

(a)          (b)          (c)          (d)

Fig. 3: **Effects of imperfect foreground segmentation** (a,b) Detected objects, as well as the foreground masks obtained from GrabCut. (c,d) Output using detection potentials. Incorrect parts of the foreground segmentation of the main aeroplane, and entire TV detection have been ignored by CRF inference as they did not agree with the other energy terms. The person is a failure case though as the detection has caused part of the sofa to be erroneously labelled.

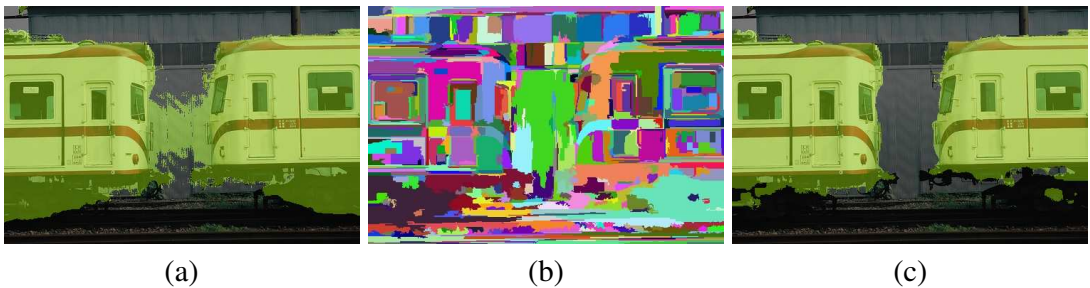

(a)                    (b)                    (c)

Fig. 4: **Segmentation enhancement from superpixel based potentials** (a) The output of our system without any superpixel potentials. (b) Superpixels obtained from the image using the method of [39]. Only one "layer" of superpixels is shown. In practice, we used four. (c) The output using superpixel potentials. The result has improved as we encourage consistency over superpixel regions. This removes some of the spurious noise that was present previously.

We use two types of energies in the CRF to encourage superpixel consistency in semantic segmentation. Firstly, we use the $P^n$-Potts model type energy [41], which is described by,

$$\psi_s^{\mathrm{SP}}(\mathbf{X}_s = \mathbf{x}_s) = \begin{cases} w_{\mathrm{Low}}(l) & \text{if all } x_s^{(i)} = l, \\ w_{\mathrm{High}} & \text{otherwise,} \end{cases} \quad (4)$$

where $w_{\mathrm{Low}}(l) < w_{\mathrm{High}}$ for all $l$, and $\{\mathbf{X}_s\} \subset \{\mathbf{X}\}$ is a clique defined by a superpixel. The primary idea is that assigning different labels to pixels in the same superpixel incurs a higher cost, whereas one obtains a lower cost if the labelling is consistent throughout the superpixel. Costs $w_{\mathrm{Low}}(l)$ and $w_{\mathrm{High}}$ are learnable during the end-to-end training of the network.

Secondly, to make this potential stronger, we average initial unary potentials from the classifier (the CNN in our case), across all pixels in the superpixel and use the average as an additional unary potential for those pixels. During experiments, we observed that superpixel based higher order energy helps in getting rid of small spurious regions of wrong labels in the segmentation output, as shown in Fig. 4.

## 5 Mean Field Updates and Their Differentials

This section discusses the mean field updates for the higher order potentials previously introduced. These update operations are differentiable with respect to the $Q_i(X_i)$ distribution inputs at each iteration, as well as the parameters of our higher order potentials. This allows us to train our CRF end-to-end as another layer of a neural network.

Take a CRF with random variables $V_1, V_2, \ldots, V_N$ and a set of cliques $\mathcal{C}$, which includes unary, pairwise and higher order cliques. Mean field inference approximates the joint distribution $\Pr(\mathbf{V} = \mathbf{v})$ with the product of marginals $\prod_i Q(V_i = v_i)$. We use $Q(\mathbf{V}_c = \mathbf{v}_c)$ to denote the marginal probability mass for a subset $\{\mathbf{V}_c\}$ of these variables. Where there is no ambiguity, we use the short-hand notation $Q(\mathbf{v}_c)$ to represent $Q(\mathbf{V}_c = \mathbf{v}_c)$. General mean field updates of such a CRF take the form [13]

$$Q^{t+1}(V_i = v) = \frac{1}{Z_i} \exp\left(-\sum_{c \in \mathcal{C}} \sum_{\{\mathbf{v}_c | v_i = v\}} Q^t(\mathbf{v}_{c-i}) \, \psi_c(\mathbf{v}_c)\right), \qquad (5)$$

where $Q^t$ is the marginal after the $t^{\text{th}}$ iteration, $\mathbf{v}_c$ an assignment to all variables in clique $c$, $\mathbf{v}_{c-i}$ an assignment to all variables in $c$ except for $V_i$, $\psi_c(\mathbf{v}_c)$ is the cost of assigning $\mathbf{v}_c$ to the clique $c$, and $Z_i$ is the normalisation constant that makes $Q(V_i = v)$ a probability mass function after the update.

**Updates from Detection Based Potentials** Following Eq. (3) above, we now use Eq. (5) to derive the mean field updates related to $\psi_d^{\text{Det}}$. The contribution from $\psi_d^{\text{Det}}$ to the update of $Q(X_d^{(i)} = l)$ takes the form

$$\sum_{\{(\mathbf{x}_d, y_d) | x_d^{(i)} = l\}} Q(\mathbf{x}_{d-i}, y_d) \, \psi_d^{\text{Det}}(\mathbf{x}_d, y_d) = \begin{cases} w_{\text{Det}} \, \frac{s_d}{n_d} \, Q(Y_d = 0) & \text{if } l = l_d, \\ w_{\text{Det}} \, \frac{s_d}{n_d} \, Q(Y_d = 1) & \text{otherwise,} \end{cases} \qquad (6)$$

where $\mathbf{x}_{d-i}$ is an assignment to $\mathbf{X}_d$ with the $i^{\text{th}}$ element deleted. Using the same equations, we derive the contribution from the energy $\psi_d^{\text{Det}}$ to the update of $Q(Y_d = b)$ to take the form

$$\sum_{\{(\mathbf{x}_d, y_d) | y_d = b\}} Q(\mathbf{x}_d) \, \psi_d^{\text{Det}}(\mathbf{x}_d, y_d) = \begin{cases} w_{\text{Det}} \, \frac{s_d}{n_d} \, \sum_{i=1}^{n_d} Q(X_d^{(i)} = l_d) & \text{if } b = 0, \\ w_{\text{Det}} \, \frac{s_d}{n_d} \, \sum_{i=1}^{n_d} (1 - Q(X_d^{(i)} = l_d)) & \text{otherwise.} \end{cases} \qquad (7)$$

It is possible to increase the number of parameters in $\psi_d^{\text{Det}}(.)$. Since we use backprop-agation to learn these parameters automatically during end-to-end training, it is desirable to have a high number of parameters to increase the flexibility of the model. Following this idea, we made the weight $w_{\text{Det}}$ class specific, that is, a function $w_{\text{Det}}(l_d)$ is used instead of $w_{\text{Det}}$ in Eqs. (2), (6) and (7). The underlying assumption is that detector outputs can be very helpful for certain classes, while being not so useful for classes that the detector performs poorly on, or classes for which foreground segmentation is often inaccurate.

Note that due to the presence of detection potentials in the CRF, error differentials calculated with respect to the $X$ variable unary potentials and pairwise parameters will no longer be valid in the forms described in [10]. The error differentials with respect to the $X$ and $Y$ variables, as well as class-specific detection potential weights $w_{\text{Det}}(l)$ are included in the supplementary material.

**Updates for Superpixel Based Potentials** The contribution from the $P^n$-Potts type potential to the mean field update of $Q(x_i = l)$, where pixel $i$ is in the superpixel clique $s$, was derived in [7] as

$$\sum_{\{\mathbf{x_s}|x_s^{(i)}=l\}} Q(\mathbf{x}_{s-i})\,\psi_s^{\text{SP}}(\mathbf{x}_s) = w_{\text{Low}}(l)\prod_{j\in c, j\neq i} Q(X_j = l) + w_{\text{High}}\left(1 - \prod_{j\in c-i} Q(X_j = l)\right). \tag{8}$$

This update operation is differentiable with respect to the parameters $w_{\text{Low}}(l)$ and $w_{\text{High}}$, allowing us to optimise them via backpropagation, and also with respect to the $Q(X)$ values enabling us to optimise previous layers in the network.

**Convergence of parallel mean field updates** Mean field with parallel updates, as proposed in [17] for speed, does not have any convergence guarantees in the general case. However, we usually empirically observed convergence with higher order potentials, without damping the mean field update as described in [7,42]. This may be explained by the fact that the unaries from the initial pixelwise-prediction part of our network provide a good initialisation. In cases where the mean field energy did not converge, we still empirically observed good final segmentations.

## 6   Experiments

We evaluate our new CRF formulation on two different datasets using the CRF-RNN network [10] as the main baseline, since we are essentially enriching the CRF model of [10]. We then present ablation studies on our models.

### 6.1   Experimental set-up and results

Our deep network consists of two conceptually different, but jointly trained stages. The first, "unary" part of our network is formed by the FCN-8s architecture [3]. It is initialised from the Imagenet-trained VGG-16 network [2], and then fine-tuned with data from the VOC 2012 training set [4], extra VOC annotations of [43] and the MS COCO [44] dataset.

Table 1: Comparison of each higher order potential with baseline on VOC 2012 reduced validation set

| Method | Reduced val set(%) |
|---|---|
| Baseline (unary + pairwise) [10] | 72.9 |
| Superpixels only | 74.0 |
| Detections only | 74.9 |
| Detections and Superpixels | 75.8 |

Table 2: Mean IoU accuracy on VOC 2012 test set. All methods are trained with MS COCO [44] data

| Method | Test set(%) |
|---|---|
| **Ours** | **77.9** |
| DPN [9] | 77.5 |
| Centrale Super Boundaries [45] | 75.7 |
| Dilated Convolutions [46] | 75.3 |
| BoxSup [35] | 75.2 |
| DeepLab Attention [47] | 75.1 |
| CRF-RNN (baseline) [10] | 74.7 |
| DeepLab WSSL [48] | 73.9 |
| DeepLab [12] | 72.7 |

Table 3: Mean Intersection over Union (IoU) results on PASCAL Context validation set compared to other current methods.

| Method | **Ours** | BoxSup [35] | ParseNet [49] | CRF-RNN [10] | FCN-8s [3] | CFM [28] |
|---|---|---|---|---|---|---|
| Mean IoU (%) | **41.3** | 40.5 | 40.4 | 39.3 | 37.8 | 34.4 |

The output of the first stage is fed into our CRF inference network. This is implemented using the mean field update operations and their differentials described in Section 5. Five iterations of mean field inference were performed during training. Our CRF network has two additional inputs in addition to segmentation unaries obtained from the FCN-8s network: data from the object detector and superpixel oversegmentations of the image.

We used the publicly available code and model of the Faster R-CNN [37] object detector. The fully automated version of GrabCut [38] was then used to obtain foregrounds from the detection bounding boxes. These choices were made after conducting preliminary experiments with alternate detection and foreground segmentation algorithms.

Four levels of superpixel oversegmentations were used, with increasing superpixel size to define the cliques used in this potential. Four levels were used since performance on the VOC validation set stopped increasing after this number. We used the superpixel method of [39] as it was shown to adhere to object boundaries the best [40], but our method generalises to any oversegmentation algorithm.

We trained the full network end-to-end, optimising the weights of the CNN classifier (FCN-8s) and CRF parameters jointly. We initialised our network using the publicly available weights of [10], and trained with a learning rate of $10^{-10}$ and momentum of 0.99. The learning rate is low because the loss was not normalised by the number of pixels in the training image. This is to have a larger loss for images with more pixels. When training our CRF, we only used VOC 2012 data [4] as it has the most accurate labelling, particularly around boundaries.

**PASCAL VOC 2012 Dataset**  The improvement obtained by each higher order potential was evaluated on the same reduced validation set [3] used by our baseline [10]. As Table 1 shows, each new higher order potential improves the mean IoU over the baseline. We only report test set results for our best method since the VOC guidelines discourage the use of the test set for ablation studies. On the test set (Table 2), we outperform our baseline by 3.2% which equates to a 12.6% reduction in the error rate. This sets a new state-of-the-art on the VOC dataset. Qualitative results highlighting success and failure cases of our algorithm, as well as more detailed results, are shown in our supplementary material.

**PASCAL Context**  Table 3 shows our state-of-the-art results on the recently released PASCAL Context dataset [50]. We trained on the provided training set of 4998 images, and evaluated on the validation set of 5105 images. This dataset augments VOC with annotations for all objects in the scene. As a result, there are 59 classes as opposed to the 20 in the VOC dataset. Many of these new labels are "stuff" classes such as "grass" and "sky". Our object detectors are therefore only trained for 20 of the 59 labels in this dataset. Nevertheless, we improve by 0.8% over the previous state-of-the-art [35] and 2% over our baseline [10].

## 6.2   Ablation Studies

We perform additional experiments to determine the errors made by our system, show the benefits of end-to-end training and compare our detection potentials to a simpler baseline. Unless otherwise stated, these experiments are performed on the VOC 2012 reduced validation set.

**Error Analysis**  To analyse the improvements made by our higher order potentials, we separately evaluate the performance on the "boundary" and "interior" regions in a similar manner to [35]. As shown in Fig. 5 c) and d), we consider a narrow band (trimap [16]) around the "void" labels annotated in the VOC 2012 reduced validation set. The mean IoU of pixels lying within this band is termed the "Boundary IoU" whilst the "Interior IoU" is evaluated outside this region.

Fig. 5 shows our results as the trimap width is varied. Adding the detection potentials improves the Interior IoU over our baseline (only pairwise potentials [10]) as the object detector may recognise objects in the image which the pixelwise classification stage of our network may have missed out. However, the detection potentials also improve the Boundary IoU for all tested trimap widths as well. Improving the recognition of pixels in the interior of an object also helps with delineating the boundaries since the strength of the pairwise potentials exerted by the $Q$ distributions at each of the correctly-detected pixels increase.

Our superpixel priors also increase the Interior IoU with respect to the baseline. Encouraging consistency over regions helps to get rid of spurious regions of wrong labels (as shown in Fig. 4). Fig. 5 suggests that most of this improvement occurs in the interior of an object. The Boundary IoU is slightly lower than the baseline, and this may be due to the fact that superpixels do not always align correctly with the edges of an object (the "boundary recall" of various superpixel methods are evaluated in [40]).

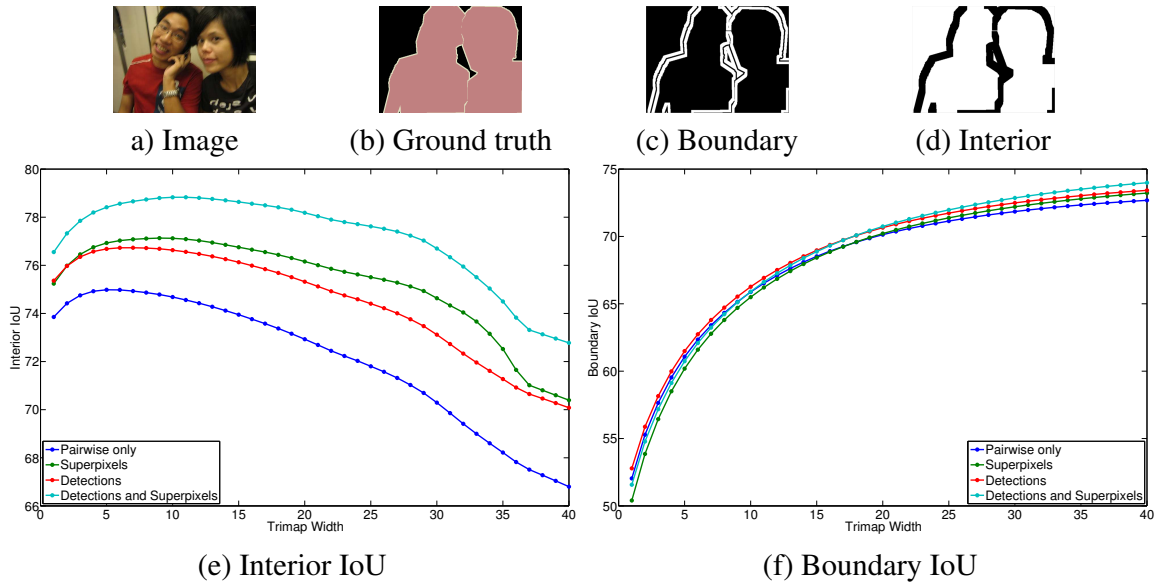

Fig. 5: **Error analysis on VOC 2012 reduced validation set** The IoU is computed for boundary and interior regions for various trimap widths. An example of the Boundary and Interior regions for a sample image using a width of 9 pixels is shown in white in the top row. Black regions are ignored in the IoU calculation.

We can see that the combination of detection and superpixel potentials results in a substantial improvement in our Interior IoU. This is the primary reason our overall IoU on the VOC benchmark increases with higher order potentials.

**Benefits of end-to-end training**   Table 4 shows how end-to-end training outperforms piecewise training. We trained the CRF piecewise by freezing the weights of the unary part of the network, and only learning the CRF parameters.

Our results in Table 2 used the FCN-8s [3] architecture to generate unaries. To show that our higher order potentials improve performance regardless of the underlying CNN used for producing unaries, we also perform an experiment using our reimplementation of the "front-end" module proposed in the Dilated Convolution Network (DCN) of [46] instead of FCN-8s.

Table 4: Comparison of mean IoU (%) obtained on VOC 2012 reduced validation set from end-to-end and piecewise training

| Method | FCN-8s | DCN |
|---|---|---|
| Unary only, fine-tuned on COCO | 68.3 | 68.6 |
| Pairwise CRF trained piecewise | 69.5 | 70.7 |
| Pairwise CRF trained end-to-end | 72.9 | 72.5 |
| Higher Order CRF trained piecewise | 73.6 | 73.5 |
| Higher Order CRF trained end-to-end | 75.8 | 75.0 |
| Test set performance of best model | 77.9 | 76.9 |

Table 4 shows that end-to-end training of the CRF yields considerable improvements over piecewise training. This was the case when using either FCN-8s or DCN for obtaining the initial unaries before performing CRF inference with higher order potentials. This suggests that our CRF network module can be plugged into different architectures and achieve performance improvements.

**Baseline for detections**  To evaluate the efficacy of our detection potentials, we formulate a simpler baseline since no other methods use detection information at inference time (BoxSup [35] derives ground truth for training using ground-truth bounding boxes).

Our baseline is similar to CRF-RNN [10], but prior to CRF inference, we take the segmentation mask from the object detection and add a unary potential proportional to the detector's confidence to the unary potentials for those pixels. We then perform mean-field inference (with only pairwise terms) on these "augmented" unaries. Using this method, the mean IoU increases from 72.9% to 73.6%, which is significantly less than the 74.9% which we obtained using only our detection potentials without superpixels (Table 1).

Our detection potentials perform better since our latent $Y$ detection variables model whether the detection hypothesis is accepted or not. Our CRF inference is able to evaluate object detection inputs in light of other potentials. Inference increases the relative score of detections which agree with the segmentation, and decreases the score of detections which do not agree with other energies in the CRF. Figures 2 b) and d) show examples of false-positive detections that have been ignored and correct detections that have been used to refine our segmentation. Our baseline, on the other hand, is far more sensitive to erroneous detections as it cannot adjust the weight given to them during inference.

## 7    Conclusion

We presented a CRF model with two different higher order potentials to tackle the semantic segmentation problem. The first potential is based on the intuitive idea that object detection can provide useful cues for semantic segmentation. Our formulation is capable of automatically rejecting false object detections that do not agree at all with the semantic segmentation. Secondly, we used a potential that encourages superpixels to have consistent labelling. These two new potentials can co-exist with the usual unary and pairwise potentials in a CRF.

Importantly, we showed that efficient mean field inference is still possible in the presence of the new higher order potentials and derived the explicit forms of the mean field updates and their differentials. This enabled us to implement the new CRF model as a stack of CNN layers and to train it end-to-end in a unified deep network with a pixelwise CNN classifier. We experimentally showed that the addition of higher order potentials results in a significant increase in semantic segmentation accuracy allowing us to reach state-of-the-art performance.

This work was supported by ERC grant ERC-2012-AdG 321162-HELIOS, EPSRC grant Seebibyte EP/M013774/1, EPSRC/MURI grant EP/N019474/1 and the Clarendon Fund.

# Appendix

Appendix A of this supplementary material presents the derivatives of the mean field updates which we use for inference in our Conditional Random Field (CRF). Appendix B shows detailed qualitative results for the experiments described in our main paper.

# A    Derivatives of Mean Field Updates

The pseudocode for the mean field inference algorithm with latent $Y$ detection variables is shown below in Algorithm 1. We use the same notation used in the main paper.

---

**Algorithm 1** Mean Field Inference

---

$Q^0(X_i = l) \leftarrow \frac{1}{Z_i} \exp\left(-\psi_i^U(l)\right), \quad \forall i, l$

$Q^0(Y_d = b) \leftarrow s_d^b (1 - s_d)^{(1-b)}, \quad \forall d, b$

$\qquad\qquad\qquad\qquad\qquad\qquad\qquad\qquad\qquad\qquad\qquad \triangleright$ Initialisation

**for** t = 0 : T − 1 **do**

$\quad E^t(X_i = l) \leftarrow \mathrm{UnaryUpdate} + \mathrm{PairwiseUpdate} +$
$\qquad\qquad\quad \mathrm{DetectionUpdate} + \mathrm{SuperpixelUpdate}, \quad \forall i, l$

$\quad E^t(Y_d = b) \leftarrow \mathrm{Y\_UnaryUpdate} + \mathrm{Y\_DetectionUpdate}$

$\qquad\qquad\qquad\qquad\qquad\qquad\qquad\qquad\qquad \triangleright$ Mean field updates

$\quad Q^{t+1}(X_i = l) \leftarrow \frac{1}{Z_i} \exp\left(-E^t(X_i = l)\right), \quad \forall i, l$

$\quad Q^{t+1}(Y_d = b) \leftarrow \frac{1}{Z_d} \exp\left(-E^t(Y_d = b)\right), \quad \forall d, b$

$\qquad\qquad\qquad\qquad\qquad\qquad\qquad\qquad\qquad\qquad\quad \triangleright$ Normalising

**end for**

---

For the explicit forms of the $\mathrm{UnaryUpdate}$ and $\mathrm{PairwiseUpdate}$ above, and their differentials, we refer the reader to [10] and discuss the terms $\mathrm{DetectionUpdate}$ and $\mathrm{SuperpixelUpdate}$ in detail below.

Let us assume that only one object detection of the form $(l_d, s_d, F_d)$ is available for the image under consideration. When multiple detections are present, simply a summation of the updates and differentials discussed below apply. Therefore, no generality is lost with this assumption. Similarly, we can assume that only one superpixel clique $\{\mathbf{X}_s\}$ is present, without a loss of generality.

Assuming that pixel $i$ in Algorithm 1 belongs to $F_d$, Eq. (6) in the main paper described the exact form of DetectionUpdate. Similarly, assuming that pixel $i$ belongs to $\{X_s\}$ Eq. (8) described the form of SuperpixelUpdate.

Let $L$ denote the value of the loss function calculated at the output of the deep network. This could be the softmax loss or any other appropriate loss function. During backpropagation, we get the error signal $\frac{\partial L}{\partial Q^T}$ at the output of the mean field inference. Using this error information, we need to compute the derivative of the loss $L$ with respect to the $X$ unaries and various CRF parameters. Note that, if we compute the relevent differentials for only one iteration of the mean field algorithm, it is possible to calculate them for multiple iterations using the recurrent behaviour of the iterations.

Note that, by looking at *Normalising* step of Algorithm 1, it is trivial to calculate $\frac{\partial Q^{t+1}}{\partial E^t}$. Therefore, we can then calculate $\frac{\partial L}{\partial E^t}$ using the chain rule. This is same as backpropagation of the usual softmax operation in a deep network (up to a negative sign). Using this observation we can calculate the necessary differentials to take the forms shown below:

$$\frac{\partial L}{\partial w_{\text{Det}}} = \frac{s_d}{n_d} \sum_{i=1}^{n_d} \Big( \frac{\partial L}{E^t(X_d^{(i)} = l_d)} Q^t(Y_d = 1) \ + \tag{9}$$
$$\sum_{l' \neq l_d} \frac{\partial L}{\partial E^t(X_d^{(i)} = l')} Q^t(Y_d = 1) \Big) \ +$$
$$\frac{\partial L}{\partial E^t(Y_d = 0)} \frac{s_d}{n_d} \sum_{i=1}^{n_d} Q^t(X_d^{(i)} = l_d) \ +$$
$$\frac{\partial L}{\partial E^t(Y_d = 1)} \frac{s_d}{n_d} \sum_{i=1}^{n_d} \Big( 1 - Q^t(X_d^{(i)} = l_d) \Big)$$

$$\frac{\partial L}{\partial Q^t(X_d^{(i)} = l_d)} = w_{\text{Det}} \frac{\partial L}{\partial E^t(Y_d = 0)} - w_{\text{Det}} \frac{\partial L}{\partial E^t(Y_d = 1)} \tag{10}$$

$$\frac{\partial L}{\partial Q^t(Y_d = 0)} = w_{\text{Det}} \frac{s_d}{n_d} \sum_{i=1}^{n_d} \left( \frac{\partial L}{E^t(X_d^{(i)} = l_d)} \right) \tag{11}$$

$$\frac{\partial L}{\partial Q^t(Y_d = 1)} = w_{\text{Det}} \frac{s_d}{n_d} \sum_{i=1}^{n_d} \sum_{l \neq l_d} \left( \frac{\partial L}{\partial E^t(X_d^{(i)} = l')} \right) \tag{12}$$

$$\frac{\partial L}{\partial w_{\text{Low}}(l)} = \sum_{i \in s} \left( \frac{\partial L}{\partial E^t(X_s^{(i)} = l)} \prod_{j \in c, j \neq i} Q^t(X_j = l) \right) \tag{13}$$

$$\frac{\partial L}{\partial w_{\text{High}}} = \sum_{i \in s} \sum_{l \in \mathcal{L}} \left( \frac{\partial L}{\partial E^t(X_s^{(i)} = l)} \left( 1 - \prod_{j \in c, j \neq i} Q^t(X_j = l) \right) \right) \qquad (14)$$

Effect of the superpixel potentials on the derivatives $\frac{\partial L}{\partial Q^t(X_i = l)}$ were negligible. Therefore, we ignored them in our calculations.

## B    Additional Experimental Results

Table 5 presents more detailed results of our method, and that of other state-of-the-art techniques, on the PASCAL VOC 2012 test set. In particular, we present the accuracy for every class in the VOC test set. Note that our per-class accuracy improves over our baseline, CRF-RNN [10], for all of the 20 classes in PASCAL VOC.

Figure 6 shows more sample results of our system, compared to our baseline, CRF-as-RNN [10]. Figure 7 shows examples of failure cases of our method. Figure 8 examines the effect of each of our potentials. Finally, Figure 9 shows a qualitative comparison between the output of our system and other current methods on the PASCAL VOC 2012 test set.

Table 5: Comparison of the mean Intersection over Union (IoU) accuracy of our approach and other state-of-the-art methods on the Pascal VOC 2012 test set. Scores for other methods were taken from the original authors' publications.

| Methods trained with COCO | Mean IoU(%) | aero-plane | bike | bird | boat | bottle | bus | car | cat | chair | cow | table | dog | horse | mbike | per-son | plant | sheep | sofa | train | tv |
|---|---|---|---|---|---|---|---|---|---|---|---|---|---|---|---|---|---|---|---|---|---|
| Our method | 77.9 | 92.5 | 59.1 | 90.3 | 70.6 | 74.4 | 92.4 | 84.1 | 88.3 | 36.8 | 85.6 | 67.1 | 85.1 | 86.9 | 88.2 | 82.6 | 62.6 | 85.0 | 56.2 | 81.9 | 72.5 |
| DPN [9] | 77.5 | 89.0 | 61.6 | 87.7 | 66.8 | 74.7 | 91.2 | 84.3 | 87.6 | 36.5 | 86.3 | 66.1 | 84.4 | 87.8 | 85.6 | 85.4 | 63.6 | 87.3 | 61.3 | 79.4 | 66.4 |
| Super Bound. [45] | 75.7 | 90.3 | 37.9 | 89.6 | 67.8 | 74.6 | 89.3 | 84.1 | 89.1 | 35.8 | 83.6 | 66.2 | 82.9 | 81.7 | 85.6 | 84.6 | 60.3 | 84.8 | 60.7 | 78.3 | 68.3 |
| Dilated Conv. [46] | 75.3 | 91.7 | 39.6 | 87.8 | 63.1 | 71.8 | 89.7 | 82.9 | 89.8 | 37.2 | 84.0 | 63.0 | 83.3 | 89.0 | 83.8 | 85.1 | 56.8 | 87.6 | 56.0 | 80.2 | 64.7 |
| BoxSup [35] | 75.2 | 89.8 | 38.0 | 89.2 | 68.9 | 68.0 | 89.6 | 83.0 | 87.7 | 34.4 | 83.6 | 67.1 | 81.5 | 83.7 | 85.2 | 83.5 | 58.6 | 84.9 | 55.8 | 81.2 | 70.7 |
| Attention [47] | 75.1 | 92.0 | 41.2 | 87.8 | 57.2 | 72.7 | 92.8 | 85.9 | 90.5 | 30.5 | 78.0 | 62.8 | 85.8 | 85.3 | 87.2 | 85.6 | 57.7 | 85.1 | 56.5 | 83.0 | 65.0 |
| CRF-as-RNN [10] | 74.7 | 90.4 | 55.3 | 88.7 | 68.4 | 69.8 | 88.3 | 82.4 | 85.1 | 32.6 | 78.5 | 64.4 | 79.6 | 81.9 | 86.4 | 81.8 | 58.6 | 82.4 | 53.5 | 77.4 | 70.1 |
| WSSL [48] | 73.9 | 89.2 | 46.7 | 88.5 | 63.5 | 68.4 | 87.0 | 81.2 | 86.3 | 32.6 | 80.7 | 62.4 | 81.0 | 81.3 | 84.3 | 82.1 | 56.2 | 84.6 | 58.2 | 76.2 | 67.2 |
| DeepLab [12] | 72.7 | 89.1 | 38.3 | 88.1 | 63.3 | 69.7 | 87.1 | 83.1 | 85.0 | 29.3 | 76.5 | 56.5 | 79.8 | 77.9 | 85.8 | 82.4 | 57.4 | 84.3 | 54.9 | 80.5 | 64.1 |
| Methods trained without COCO | | | | | | | | | | | | | | | | | | | | | |
| Our method | 73.9 | 89.3 | 40.0 | 81.6 | 65.1 | 71.7 | 90.1 | 81.3 | 85.7 | 32.4 | 82.1 | 62.2 | 82.6 | 83.7 | 84.5 | 81.1 | 60.8 | 85.2 | 49.6 | 80.0 | 69.9 |
| DPN [9] | 74.1 | 87.7 | 59.4 | 78.4 | 64.9 | 70.3 | 89.3 | 83.5 | 86.1 | 31.7 | 79.9 | 62.6 | 81.9 | 80.0 | 83.5 | 82.3 | 60.5 | 83.2 | 53.4 | 77.9 | 65.0 |
| DeconvNet [51] | 72.5 | 89.9 | 39.3 | 79.7 | 63.9 | 68.2 | 87.4 | 81.2 | 86.1 | 28.5 | 77.0 | 62.0 | 79.0 | 80.3 | 83.6 | 80.2 | 58.8 | 83.4 | 54.3 | 80.7 | 65.0 |
| CRF-as-RNN [10] | 72.0 | 87.5 | 39.0 | 79.7 | 64.2 | 68.3 | 87.6 | 80.8 | 84.4 | 30.4 | 78.2 | 60.4 | 80.5 | 77.8 | 83.1 | 80.6 | 59.5 | 82.8 | 47.8 | 78.3 | 67.1 |
| DeepLab [12] | 71.6 | 84.4 | 54.5 | 81.5 | 63.6 | 65.9 | 85.1 | 79.1 | 83.4 | 30.7 | 74.1 | 59.8 | 79.0 | 76.1 | 83.2 | 80.8 | 59.7 | 82.2 | 50.4 | 73.1 | 63.7 |
| Piecewise [11] | 70.7 | 87.5 | 37.7 | 75.8 | 57.4 | 72.3 | 88.4 | 82.6 | 80.0 | 33.4 | 71.5 | 55.0 | 79.3 | 78.4 | 81.3 | 82.7 | 56.1 | 79.8 | 48.6 | 77.1 | 66.3 |
| Zoomout [52] | 69.6 | 85.6 | 37.3 | 83.2 | 62.5 | 66.0 | 85.1 | 80.7 | 84.9 | 27.2 | 73.2 | 57.5 | 78.1 | 79.2 | 81.1 | 77.1 | 53.6 | 74.0 | 49.2 | 71.7 | 63.3 |
| FCN-8s [3] | 62.2 | 76.8 | 34.2 | 68.9 | 49.4 | 60.3 | 75.3 | 74.7 | 77.6 | 21.4 | 62.5 | 46.8 | 71.8 | 63.9 | 76.5 | 73.9 | 45.2 | 72.4 | 37.4 | 70.9 | 55.1 |
| CFM [28] | 61.8 | 75.7 | 26.7 | 69.5 | 48.8 | 65.6 | 81.0 | 69.2 | 73.3 | 30.0 | 68.7 | 51.5 | 69.1 | 68.1 | 71.7 | 67.5 | 50.4 | 66.5 | 44.4 | 58.9 | 53.5 |
| NUS_UDS [53] | 50.0 | 67.0 | 24.5 | 47.2 | 45.0 | 47.9 | 65.3 | 60.6 | 58.5 | 15.5 | 50.8 | 37.4 | 45.8 | 59.9 | 62.0 | 52.7 | 40.8 | 48.2 | 36.8 | 53.1 | 45.6 |
| O2P [54] | 47.8 | 64.0 | 27.3 | 54.1 | 39.2 | 48.7 | 56.6 | 57.7 | 52.5 | 14.2 | 54.8 | 29.6 | 42.2 | 58.0 | 54.8 | 50.2 | 36.6 | 58.6 | 31.6 | 48.4 | 38.6 |

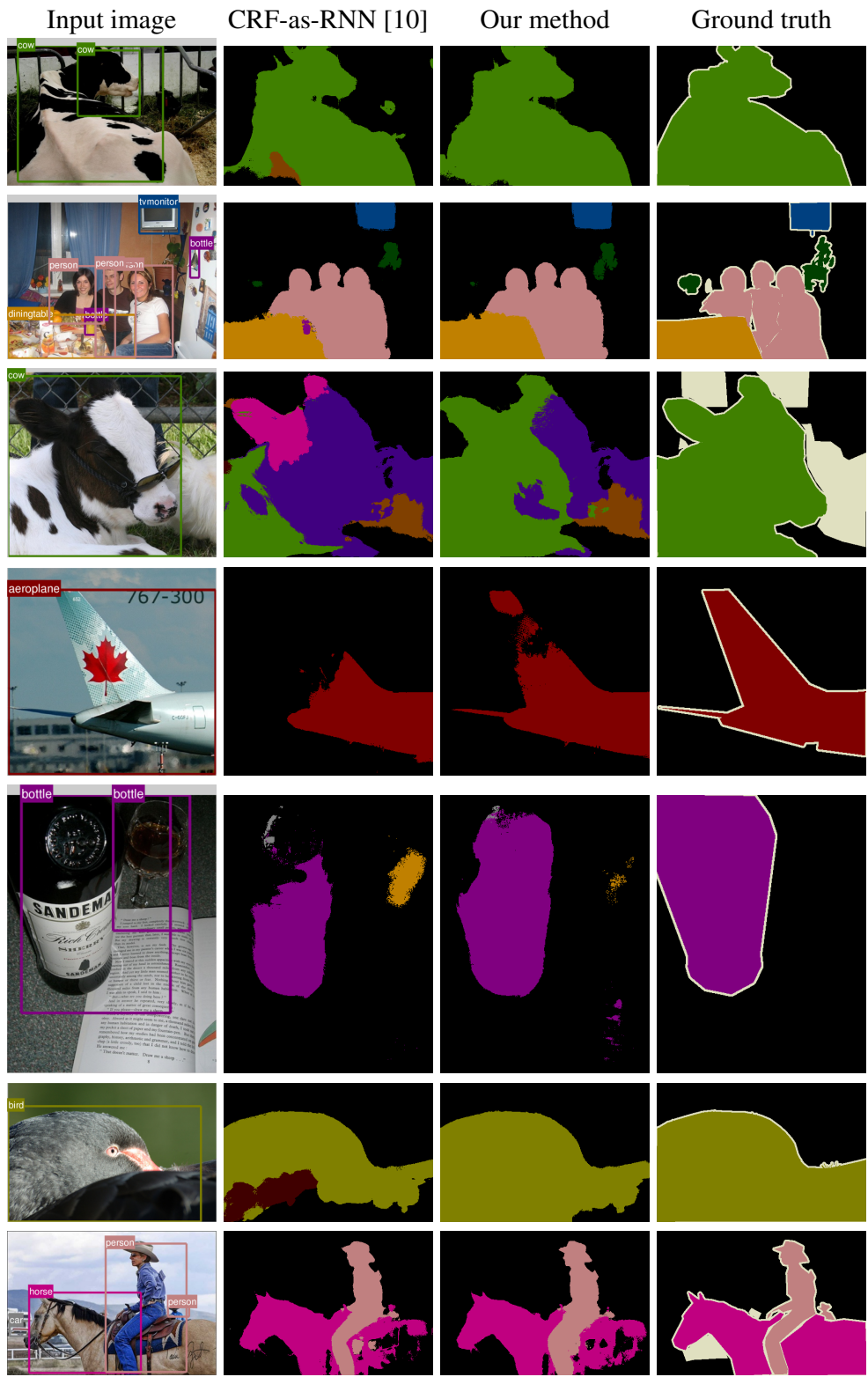

Fig. 6: **Examples of images where our method has improved over our baseline, CRF-as-RNN [10].** The input images have the detection bounding boxes overlaid on them. Note that the method of [10] does not make use of this information. The improvements from our method are due to our detection potentials, as well as our superpixel based potentials. Note that all images are from the reduced validation set of VOC 2012 and have not been trained on at all. Best viewed in color.

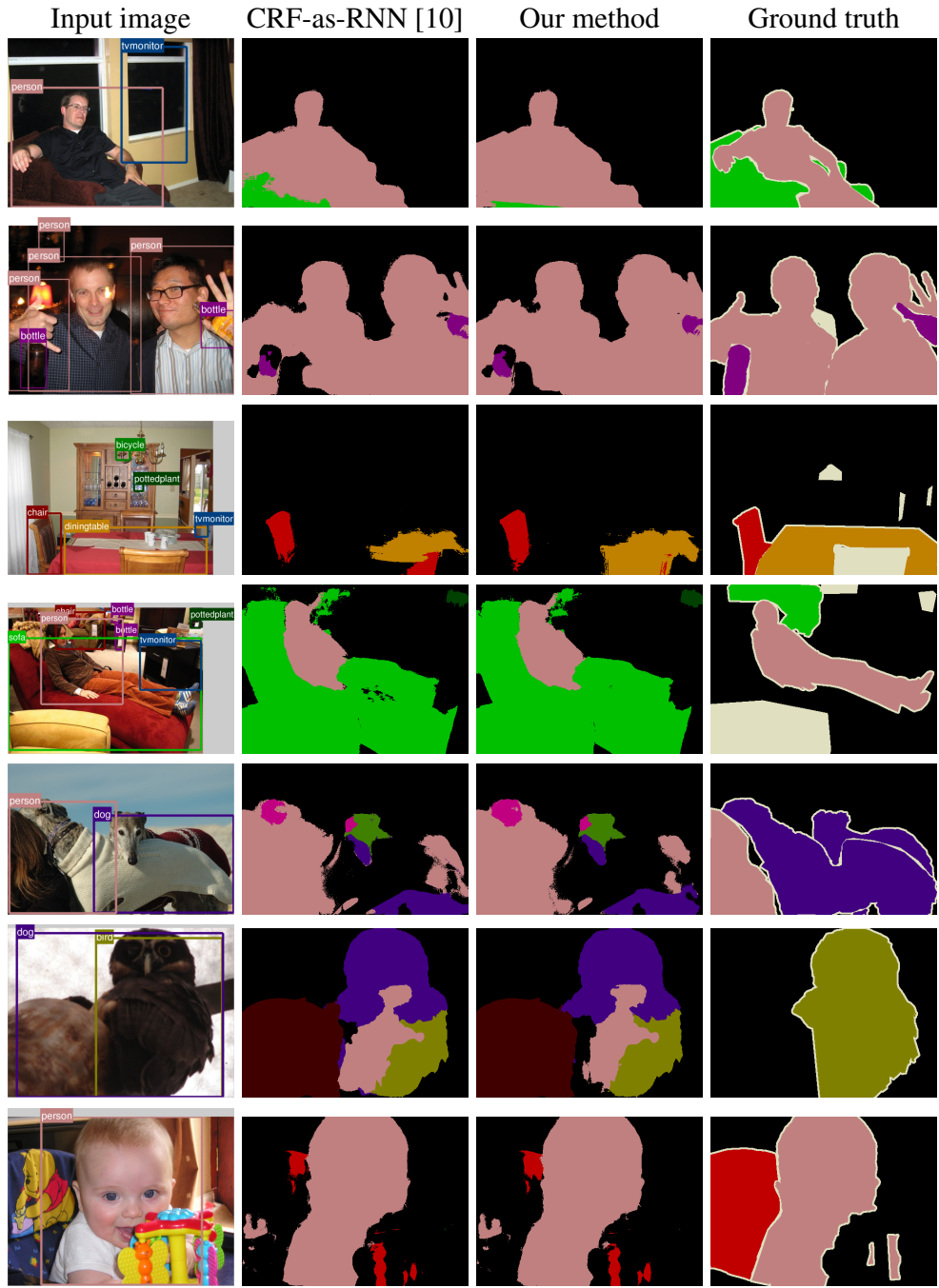

Fig. 7: **Examples of failure cases where our method has performed poorly.** The first row shows an example of how the detection of the person has now resulted in the sofa being misclassified (although our system is able to reject the other false detection). Our superpixel potentials have a tendency to remove spurious noise by enforcing consistency within regions. However, as shown in the second row, sometimes the "noise" being removed is actually the correct label. In the other cases, we are limited by our pixelwise classification unaries which are poor. Our superpixel and detection potentials are not always able to compensate for this. Note that all images are from the reduced validation set of VOC 2012 and have not been trained on at all. The input images have the detection bounding boxes overlaid on them. Note that the method of [10] does not make use of this information. Best viewed in color.

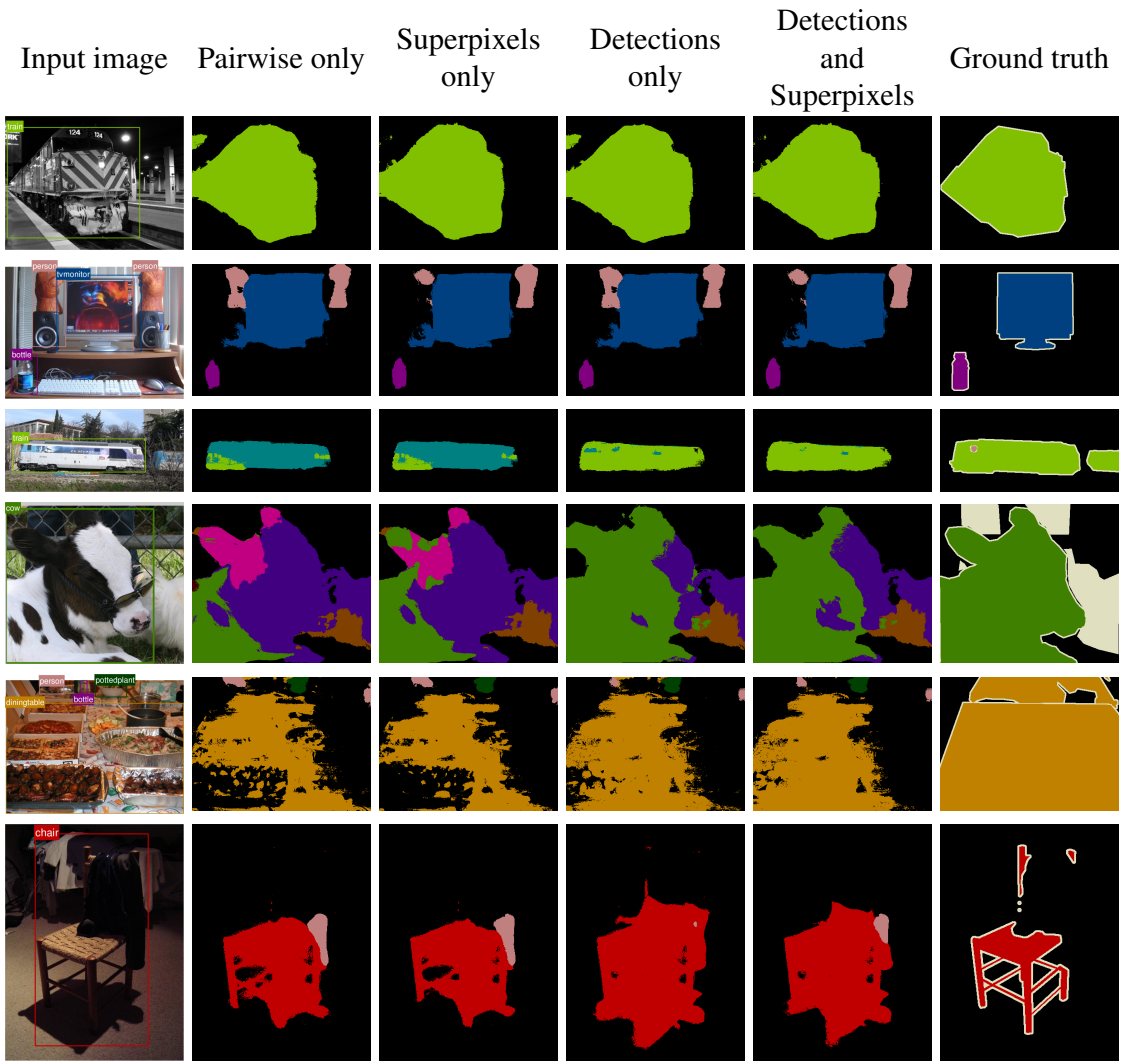

Fig. 8: **Comparison of pairwise potentials, superpixel and pairwise potentials, detection and pairwise potentials, and a combination of all three** (Row 1 and 2) These are examples where superpixel potentials help to remove spurious noise in the output but detection potentials do not affect the result. The final result still improves when all potentials are combined. (Row 3) Detection potentials greatly improve the result by recognising the train correctly (the pixelwise unaries are largest for "bus"). And superpixels, when combined with detections, slightly improve the output. (Row 4) An example where both superpixel and detection potentials improve the final output. (Row 5) A case where the superpixel worsens the result as, although the output is more consistent among superpixel regions, some pixels have had their correct labels removed. However, the correct detection improves the result, and the output of combining superpixel and detection potentials is actually better than either potential in isolation. (Row 6) Here, the detection (although correct) worsens the output due to its imprecise foreground mask. Superpixel potentials also exacerbate the result, since the legs of the chair and the chair's shadow are confused to be part of the same superpixel region. However, when the two potentials are combined, the result is slightly better than with only detection potentials.

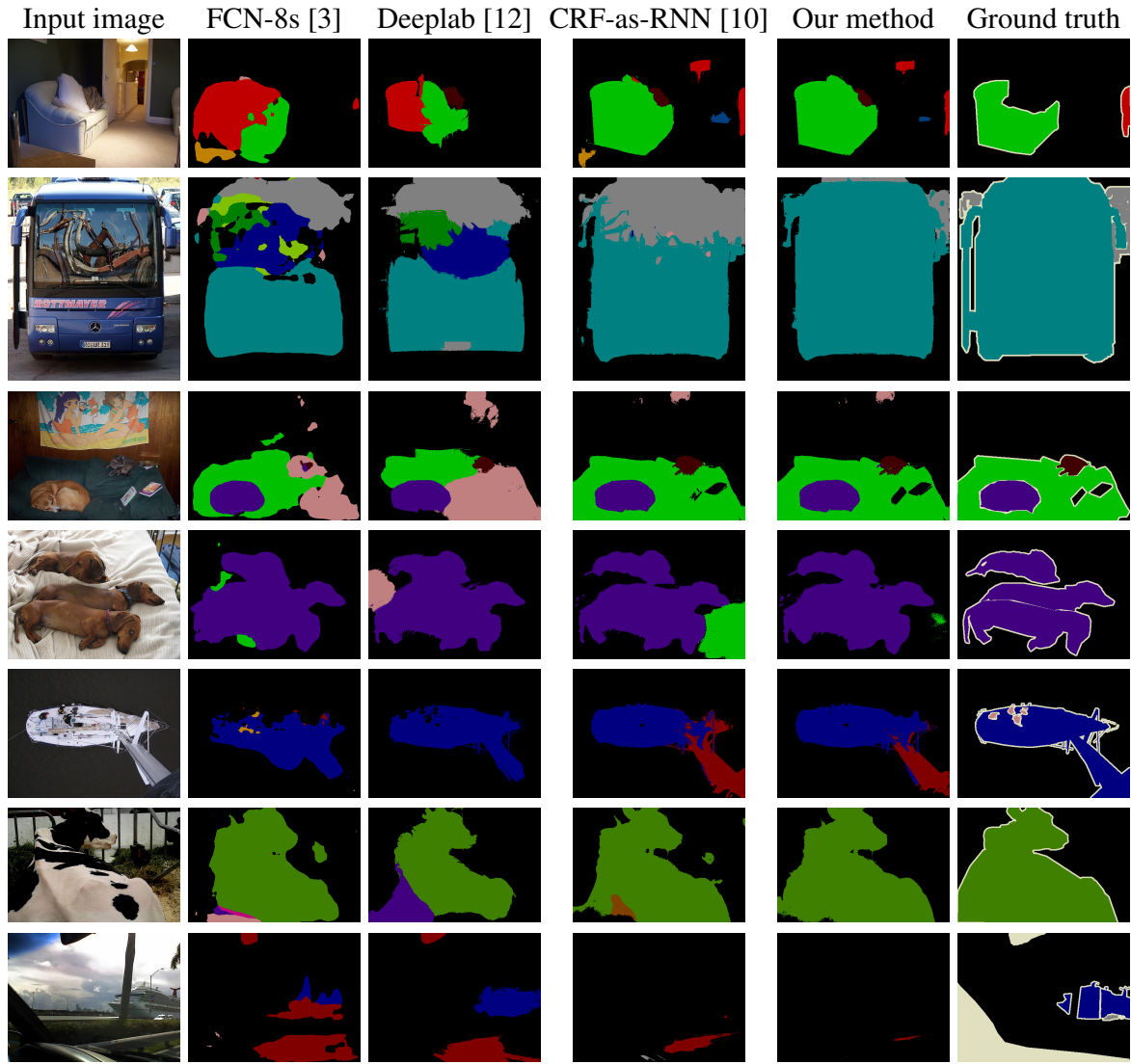

Fig. 9: **Qualitative comparison with other current methods.** Sample results of our method compared to other current techniques on VOC 2012. We reproduced the segmentation results of Deeplab from their original publication, whilst we reproduced the results of FCN-8s and CRF-as-RNN from their publicly-available source code. Best viewed in colour.